\documentclass[11pt]{article}
\usepackage[margin=1in]{geometry}
\usepackage{amsmath,amssymb,amsthm,mathtools}
\usepackage[T1]{fontenc}
\usepackage{lmodern,microtype}
\usepackage[hidelinks]{hyperref}
\hypersetup{pdftitle={Proper Agnostic Learning of Affine Halfspaces under Log-Concave Marginals},pdfauthor={Zhijie Zheng}}
\newtheorem{theorem}{Theorem}[section]
\newtheorem{lemma}[theorem]{Lemma}
\newtheorem{proposition}[theorem]{Proposition}
\newtheorem{corollary}[theorem]{Corollary}
\theoremstyle{remark}\newtheorem{remark}[theorem]{Remark}
\newcommand{\R}{\mathbb R}\newcommand{\E}{\mathbb E}\newcommand{\Pp}{\mathbb P}
\newcommand{\norm}[1]{\left\lVert#1\right\rVert}\newcommand{\abs}[1]{\left|#1\right|}
\newcommand{\Cov}{\operatorname{Cov}}\newcommand{\Var}{\operatorname{Var}}\newcommand{\tr}{\operatorname{tr}}
\newcommand{\poly}{\operatorname{poly}}\newcommand{\OPT}{\operatorname{OPT}}
\newcommand{\ind}{\mathbf 1}\newcommand{\err}{\operatorname{err}}\newcommand{\sos}{\preceq_{\rm sos}}
\title{Proper Agnostic Learning of Affine Halfspaces\linebreak under Log-Concave Marginals}
\author{Zhijie Zheng\\Peking University\\\texttt{2501210512@stu.pku.edu.cn}\\\small\href{https://orcid.org/0009-0006-7610-0691}{ORCID: 0009-0006-7610-0691}}
\date{}
\begin{document}\maketitle
\begin{abstract}
We give a proper constant-factor agnostic learner for affine halfspaces under full-dimensional log-concave marginals. For arbitrary binary labels, the algorithm uses independent labeled samples and returns a single affine halfspace with error at most $C\OPT+\varepsilon$. Its sample and real-arithmetic complexity are polynomial in dimension, inverse accuracy, and the logarithm of the reciprocal failure probability. The mean and covariance of the marginal need not be known, and the sample complexity can be made linear in this logarithmic confidence factor. The main geometric result compares global disagreement with disagreement on a balanced restriction: for a minority halfspace of mass $p$, global disagreement is within universal constant factors of $p$ times conditional disagreement, provided the latter is sufficiently small. To construct the restriction from noisy labels, we estimate the minority cap's mean and covariance robustly, approximately minimize the classical zonoid-depth dual objective, and expand a superlevel set to a marginal quantile. The disagreement comparison cancels the noise amplification caused by conditioning and allows the learned halfspace to be used on the entire space. The analysis combines degree-four sum-of-squares moment bounds for log-concave laws with a translation-based reduction from balanced affine learning to homogeneous halfspace learning.
\end{abstract}

\section{Introduction}
Learning a halfspace with arbitrary adversarial label noise is a basic problem in computational learning theory. Under isotropic log-concave marginals, constant-factor agnostic learning is known for origin-centered halfspaces through localization and nonconvex optimization \cite{abl17,dktz20}. Arbitrary affine offsets create a different difficulty: a target can describe a very rare tail, and an error bound relative to the conditional tail law need not control the original risk.

For affine halfspaces under general isotropic log-concave marginals, Diakonikolas, Kane, and Stewart \cite{dks18} give a proper learner with square-root-type dependence on the optimal error. Under Gaussian marginals, constant-factor guarantees are known for arbitrary affine halfspaces \cite{dks18,dktz22}. We obtain a constant-factor guarantee for all full-dimensional log-concave marginals, with complexity jointly polynomial in dimension and inverse accuracy.

Write $P$ for the marginal law of $X$, and let $\mathcal H_d$ be the class consisting of all affine halfspaces and the two constants. For a joint law $D$ of $(X,Y)$ with $Y\in\{0,1\}$, set
\[
 \err_D(h)=\Pp_D(h(X)\ne Y),\qquad
 \OPT_D=\inf_{h\in\mathcal H_d}\err_D(h).
\]
A full-dimensional law is log-concave if it has a density whose logarithm is concave on its convex support; zero values outside the support are allowed. Isotropy means $\E X=0$ and $\Cov(X)=I$.

\begin{theorem}[Isotropic log-concave marginals]\label{thm:main}
There are a universal constant $C$ and a randomized algorithm with the following guarantee. For every joint law $D$ with isotropic log-concave marginal $P$ on $\R^d$, every $\varepsilon,\delta\in(0,1/2)$, and arbitrary binary labels, the algorithm uses iid labeled samples alone and returns $\widehat h\in\mathcal H_d$ such that
\[
 \err_D(\widehat h)\le C\OPT_D+\varepsilon
\]
with probability at least $1-\delta$. Its sample count and real-arithmetic operation count are
\[
 \poly(d,1/\varepsilon,\log(1/\delta)).
\]
\end{theorem}

The learner uses ordinary independent samples; each conditional sample is obtained by rejection from these samples. It requires neither the marginal density nor the comparator's parameters, error, or minority mass. The marginal may have a nonsmooth density or a compact convex support.

\begin{corollary}[Unknown mean and covariance]\label{cor:general}
There are a universal constant $C$, a fixed polynomial $P$, and a randomized algorithm such that, for every joint law $D$ with a full-dimensional log-concave marginal on $\R^d$ and binary labels, and every $\varepsilon,\delta\in(0,1/2)$, the algorithm returns
\[
 \widehat h(x)=\ind\{w^Tx\ge b\},\qquad w\ne0,\quad b\in\R,
\]
with $\err_D(\widehat h)\le C\OPT_D+\varepsilon$ with probability at least $1-\delta$. It uses at most
\[
 P(d,\varepsilon^{-1})\log(2/\delta)
\]
independent labeled samples and $\poly(d,\varepsilon^{-1},\log(2/\delta))$ real-arithmetic operations. The mean and covariance are unknown, and these bounds are independent of the covariance condition number.
\end{corollary}

Section~\ref{sec:affine-extension} proves the corollary by empirical whitening and confidence amplification. It also replaces constant candidates by finite-threshold halfspaces. Theorem~\ref{thm:main} is stated in isotropic coordinates to simplify the geometric argument.

\paragraph{Computational model.}
We count real-arithmetic operations, comparisons, square roots, matrix decompositions, and the operations of the fixed-degree semidefinite procedures used for robust moment estimation. All sample requests and optimization stages have predetermined polynomial budgets. Section~\ref{sec:implementation} implements the sigmoid gradient to sufficient accuracy using arithmetic operations, so an exact exponential oracle is unnecessary. The bounds concern this real-arithmetic model; they do not bound the bit lengths of intermediate quantities.

\paragraph{Proof overview and related work.}

Let $H$ be a minority halfspace of mass $p$, and let $Q=P\mid H$. In coordinates centered at the mean of $Q$ and normalized by its covariance, $Q$ is isotropic and the target has a bounded offset. In these coordinates the convex objective $v\mapsto\E_P(1+v^TZ)_+$ has minimum $p$ at the normalized target. This objective is the classical zonoid-depth dual, related to the cap-barycenter representations of \cite{km97,kt12}. Robust estimates of the cap moments yield approximate coordinates in which the objective can be minimized to additive $O(p)$. We then expand a central superlevel set to a quantile restriction of mass $\Theta(p)$. The restriction balances $H$ and has covariance comparable to $Q$.

A balanced affine target is reduced to homogeneous learning by translating along the segment between its two conditional class means. Same-label means and a grid of centers already appear in Gaussian testable affine learning \cite{dklz24}; our reduction verifies the needed two-dimensional density conditions for general log-concave marginals. The main geometric transfer is that a sufficiently accurate affine classifier on the balanced restriction has global disagreement at most $Cp$ times its conditional disagreement. The factor $p$ cancels the $1/p$ increase in label noise. The resulting affine halfspace is used on the whole space.

We use Letwin's quadratic-form variance theorem \cite{letwin26} to obtain a universal fourth-degree sum-of-squares certificate, also recorded in \cite{storozhenko26}. Kothari--Steurer's robust moment estimator \cite{ks17} then supplies covariance estimates of constant relative accuracy and a small constant Mahalanobis error for the mean. Only fourth-moment certifiability and fixed estimation accuracy are required. The homogeneous learning step extends the geometric argument of \cite{dktz20} to the translated marginals arising in this reduction.

\section{Elementary log-concave geometry}\label{sec:elementary}
Constants $c,C>0$ below are universal, unless their fixed dependencies are displayed. They may change from line to line. Linear marginals and positive-probability convex restrictions of log-concave laws are log-concave. We use the following standard fixed-dimensional facts \cite{lv07,bl19}. For $j\in\{1,2\}$, an isotropic log-concave density $f$ on $\R^j$ satisfies
\begin{equation}\label{eq:core}
 f(z)\le A e^{-a\norm z},\qquad f(z)\ge a\quad(\norm z\le r)
\end{equation}
for absolute $a,A,r>0$. If instead $c_0I\preceq\Cov\preceq C_0I$, the same conclusions hold with adjusted constants and with the lower ball centered at the mean. A bounded mean also yields an exponential upper envelope about zero. One-dimensional log-concave densities of standard deviation $\sigma$ obey
\begin{equation}\label{eq:density-sd}
 c\le \sigma\sup f\le C.
\end{equation}
Their centered moments of every fixed order are bounded after variance normalization. We use upper-semicontinuous log-concave versions of densities; boundary identities may equivalently be read as traces from the interior.

\begin{lemma}[Restrictions of constant probability]\label{lem:restriction}
Let $Z$ be isotropic log-concave and let $A$ be a convex event with $\Pp(A)\ge a_0>0$. Then
\[
 \norm{\E[Z\mid A]}\le C(a_0),\qquad
 c(a_0)I\preceq\Cov(Z\mid A)\preceq C(a_0)I.
\]
\end{lemma}
\begin{proof}
For every unit $v$, the conditional second moment is at most $1/a_0$. This bounds the covariance above and, by taking a directional supremum, bounds the norm of the conditional mean. The conditional density of $v^TZ$ is pointwise at most the unconditional density divided by $a_0$, hence at most $C/a_0$. Any scalar density bounded by $B$ has variance at least $c/B^2$: for every center $t$, the interval of radius $1/(4B)$ contains at most half the mass. This proves the lower bound. Convexity preserves log-concavity for subsequent applications.
\end{proof}

\begin{lemma}[Interval mass]\label{lem:interval}
If a one-dimensional log-concave law gives probability at least $a_0$ to $[-R,R]$, then its variance and squared mean are at most $C(a_0,R)$.
\end{lemma}
\begin{proof}
The density supremum is at least $a_0/(2R)$. Equation~\eqref{eq:density-sd} bounds its standard deviation by $2CR/a_0$. If its mean is outside the interval, Chebyshev gives
$a_0\le\sigma^2/(\abs\mu-R)^2$, which also bounds the mean.
\end{proof}

For a centered log-concave law, every halfspace through its mean has probability at least $1/e$ \cite{lv07}. This also follows by the one-dimensional form of the log-concave centroid inequality. It applies in any dimension by projection.

\section{Robust cap statistics at a fixed degree}\label{sec:robust}
We use robust mean and covariance estimates with fixed relative accuracy. This permits a sum-of-squares algorithm of fixed degree throughout the proof.

\begin{lemma}[Fixed-accuracy robust log-concave moments]\label{lem:robust}
Fix $r\in(0,1/4)$. There is a constant $\gamma_r>0$ and a polynomial-time algorithm such that, from a sufficiently large polynomial-size sample from a full-dimensional log-concave law $Q$ with mean $\mu$ and covariance $\Sigma$, after at most a $\gamma_r$ fraction of arbitrary replacements, it returns estimates obeying
\begin{equation}\label{eq:robust}
 (1-r)\Sigma\preceq\widehat\Sigma\preceq(1+r)\Sigma,
 \qquad \norm{\Sigma^{-1/2}(\widehat\mu-\mu)}\le r
\end{equation}
with probability at least a fixed constant greater than $1/2$. Neither $\mu$ nor $\Sigma$ is supplied. Independent repetitions may be retained as a list to obtain high confidence that at least one output obeys~\eqref{eq:robust}.
\end{lemma}
\begin{proof}
Let $Z=\Sigma^{-1/2}(X-\mu)$. Letwin's Theorem 1.2 \cite{letwin26} states, for every symmetric $M$,
\begin{equation}\label{eq:letwin}
 \Var(Z^TMZ)\le 2\E\norm{\nabla(Z^TMZ)}^2=8\norm M_F^2.
\end{equation}
The difference between the two sides is a positive-semidefinite quadratic form in the entries of $M$. Substituting $M=vv^T$ therefore gives a degree-four sum-of-squares certificate for
\begin{equation}\label{eq:certificate}
 \E(v^TZ)^4\sos 9\norm v^4.
\end{equation}
Indeed the variance subtracts $(\tr M)^2$, and both $\norm{vv^T}_F^2$ and $(\tr(vv^T))^2$ equal $\norm v^4$. This certificate also appears in \cite[Lemma 3.2]{storozhenko26}. Affine changes give the corresponding relative certificate with $(v^T\Sigma v)^2$ on the right.

Theorem 1.2 of Kothari--Steurer \cite{ks17}, applied with $k=4$, gives Mahalanobis mean error $O(\gamma^{3/4})$ and relative Loewner covariance error $O(\sqrt\gamma)$ for a universal certificate constant. Taking $\gamma_r$ sufficiently small gives~\eqref{eq:robust}. The degree, and hence the polynomial exponent, remains fixed.

For completeness, the clean empirical distribution meets the required certificate with fixed slack. On the space of symmetric matrices, let $T=\E[\operatorname{vec}(ZZ^T)\operatorname{vec}(ZZ^T)^T]$ and let $i=\operatorname{vec}(I)$. Equation~\eqref{eq:letwin} gives $8I+ii^T-T\succeq0$. Also
\[
 \E\norm Z^8\le C d^4,
\]
by the scalar eighth-moment bound and $(\sum_j Z_j^2)^4\le d^3\sum_jZ_j^8$. Thus the empirical tensor $T_n$ satisfies
\[
 \E\norm{T_n-T}_F^2\le Cd^4/n.
\]
Adding fixed identity slack to the Gram matrix before sampling gives a raw empirical quartic certificate with high constant probability for polynomial $n$. Empirical first and second moments concentrate with polynomial samples as well. The degree-four sum-of-squares inequality $(s-t)^4\le8(s^4+t^4)$ transfers the certificate to empirical centering, and an empirical covariance lower bound converts $\norm v^4$ to a fixed multiple of the squared empirical variance. Hence a universal relative empirical certificate is available. No degree-eight certificate is required: ordinary eighth moments are used only for concentration. Applying the robust theorem to this empirical law, and absorbing clean sampling errors by slightly tighter constants, proves the lemma.
\end{proof}

\begin{remark}
Positive-probability halfspace restrictions remain log-concave. Consequently, the same degree-four certificate and robust estimator apply to every clean conditional law used below.
\end{remark}

\section{Minority-cap normalization and a convex restriction}\label{sec:cap}
Let $P$ be a full-dimensional log-concave law, $H$ a halfspace with mass $p\in(0,1/2]$, and $Q=P\mid H$. Write $\mu=\E_QX$, $\Sigma=\Cov_QX$, and use the exact coordinates
\begin{equation}\label{eq:coordinates}
 Z=\Sigma^{-1/2}(X-\mu).
\end{equation}
The law of $Z$ under $Q$ is isotropic. Positive-probability halfspace restrictions have positive-definite covariance.

\begin{lemma}[Normalized cap offset]\label{lem:offset}
In coordinates~\eqref{eq:coordinates}, write
\[
 H=\{n^TZ+a\ge0\},\qquad \norm n=1.
\]
There is an absolute $A_0$ such that $1\le a\le A_0$.
\end{lemma}
\begin{proof}
Under $Q$, the nonnegative residual $W=u^TX-b$ of the original defining halfspace is log-concave. Its survival function $S$ is log-concave with $S(0)=1$. Concavity of $\log S$ implies $S(s+t)\le S(s)S(t)$ for $s,t\ge0$. Integrating gives
\[
 (\E W)^2\ge\int_0^\infty\!\int_0^\infty S(s+t)\,ds\,dt
 =\tfrac12\E W^2.
\]
Thus $\Var W\le(\E W)^2$, so $a=\E W/\sqrt{\Var W}\ge1$.

For the upper bound, let $q$ be the density of $n^TZ$ under $Q$, and let $f$ be its original marginal density divided by $p$. On $[-a,\infty)$, $f=q$, while
\[
 \int_{-\infty}^{-a}f(x)\,dx=(1-p)/p\ge1.
\]
By~\eqref{eq:core}, $q(0)\ge c$ and $q(-a)\le Ae^{-ca}$. For large $a$, the secant slope
\[
 s=\frac{\log q(0)-\log q(-a)}{a}
\]
is at least a fixed positive constant. Concavity of $\log f$ bounds its left extrapolation by
\[
 f(x)\le q(-a)e^{s(x+a)}\qquad(x\le-a).
\]
The outside integral is at most $q(-a)/s<1$ when $a$ is sufficiently large, a contradiction. If a boundary density vanishes, a positive value both to its left and at zero would contradict log-concavity. Taking interior traces therefore covers that case as well.
\end{proof}

\begin{lemma}[The zonoid-depth dual objective]\label{lem:hinge}
In exact cap coordinates,
\[
 L(v)=\E_P(1+v^TZ)_+
\]
has a global minimizer $v_*=n/a$, with $\norm{v_*}\le1$ and $L(v_*)=p$. More precisely,
\begin{equation}\label{eq:hingeidentity}
 L(v)-p=\E_P\!\left[\abs{1+v^TZ}\,
 \ind_{H\triangle\{1+v^TZ\ge0\}}\right].
\end{equation}
If~\eqref{eq:robust} holds with a sufficiently small fixed $r$, then in the computable coordinates $\widehat Z=\widehat\Sigma^{-1/2}(X-\widehat\mu)$ the true halfspace is $\{1+\widehat v_*^T\widehat Z\ge0\}$ with $\norm{\widehat v_*}\le B$ and
\[
 \E_P(1+\widehat v_*^T\widehat Z)_+\le C_0p,
\]
for universal constants $B,C_0$.
\end{lemma}
\begin{proof}
The exact hinge vanishes off $H$, and its expectation under $Q$ is one. Also $\E_P[Z\ind_H]=0$, so the hinge's gradient at $v_*$ vanishes. Subtracting this supporting tangent of the positive-part function gives~\eqref{eq:hingeidentity} pointwise after integration.

For the approximate assertion, write $Z=A\widehat Z+e$, where $A=\Sigma^{-1/2}\widehat\Sigma^{1/2}$ and $e=\Sigma^{-1/2}(\widehat\mu-\mu)$. The true score is $n^TA\widehat Z+(a+n^Te)$. Its intercept $c_*=a+n^Te$ is bounded below by $1-r$, while $\norm{A^Tn}$ is bounded above by a universal constant. Dividing by $c_*$ gives the required coefficient bound. The original expectation of the positive score is $pa$, so the normalized hinge loss is $pa/c_*\le C_0p$.
\end{proof}

The objective in Lemma~\ref{lem:hinge} has a classical interpretation. The optimization
\[
 \sup\{\E_Pg:0\le g\le1,\ \E_P[gZ]=0\}
\]
defines zonoid depth at the cap mean. Its formal Lagrange dual is $\inf_vL(v)$. Here $g=\ind_H$ and $v=n/a$ explicitly attain matching values $p$, establishing equality of the primal and dual optima. Zonoid trimming and halfspace barycentric representation are established in \cite{km97,kt12}; we use approximate cap statistics to construct the balanced restriction below.

\begin{proposition}[Balancing by a quantile restriction]\label{prop:balance}
Suppose estimates obey~\eqref{eq:robust}, and find $v$ with $\norm v\le B$ and
\[
 \E_P(1+v^T\widehat Z)_+\le C_1p.
\]
For a sufficiently large fixed $K_0$, if $p$ is sufficiently small, one can choose a halfspace $K$ containing
\[
 S=\{v^T\widehat Z\ge-1/2\}
\]
with $P(K)\in[K_0p/2,2K_0p]$. For $R=P\mid K$, in the exact cap coordinates~\eqref{eq:coordinates},
\begin{align}
 c/K_0&\le R(H)\le2/K_0,\label{eq:balance}\\
 \norm{\E_RZ}&\le C,\qquad cI\preceq\Cov_RZ\preceq CI.\label{eq:regioncov}
\end{align}
The constants in~\eqref{eq:regioncov} depend only on the fixed $K_0$. All these conclusions are stable under fixed-factor errors in the target mass $K_0p$ and a sufficiently accurate empirical quantile.
\end{proposition}
\begin{proof}
The hinge is at least $1/2$ on $S$, so $P(S)\le2C_1p$. Under $Q$, the approximate coordinate mean has norm $O(r)$. Take $r$ small enough that $\abs{v^T\E_Q\widehat Z}\le1/4$ for every feasible $v$. Then $S$ contains the halfspace $\{v^T\widehat Z\ge v^T\E_Q\widehat Z\}$, and the centered log-concave halfspace bound gives $Q(S)\ge1/e$. For $p$ in the rare regime, $v=0$ is impossible because its hinge loss is one. Its scalar marginal is atomless, so lowering the threshold expands $S$ to any specified larger probability below one. Choose $K_0$ with fixed slack beyond $2C_1$.

Put $q_K=Q(K)\ge1/e$. Then
\[
 R(H)=\frac{p q_K}{P(K)},
\]
which gives~\eqref{eq:balance}. The law $Q_K=Q\mid K$ has bounded mean and covariance comparable to $I$ by Lemma~\ref{lem:restriction}. It is a component of $R$ with weight $p q_K/P(K)\ge c'>0$. Covariance decomposition gives the lower bound in~\eqref{eq:regioncov}.

For the upper bound, fix any unit $w$. The law $Q_K$ puts a fixed positive mass in $\abs{w^TZ}\le C$ by its bounded directional second moment. So does $R$, by the same component weight. The $w$ marginal of $R$ is log-concave, since $K$ is a halfspace. Lemma~\ref{lem:interval} bounds its variance and squared mean by a fixed constant. Taking suprema over unit $w$ proves both remaining assertions. 

Finally, all probability inequalities have fixed slack. A quantile with additive error a sufficiently small multiple of $p$ still expands $S$ and has the required mass bounds, after increasing $K_0$ if necessary.
\end{proof}

\section{A homogeneous primitive for translated marginals}\label{sec:homogeneous}
The homogeneous learner will be applied after a translation. We therefore formulate its assumptions directly in terms of two-dimensional marginal densities.

\begin{lemma}[Translated homogeneous learning]\label{lem:homogeneous}
Let a marginal $M$ on $\R^d$, $d\ge2$, have every two-dimensional marginal density $f_E$ satisfying
\begin{equation}\label{eq:wellbehaved}
 f_E(z)\ge c_0\quad(\norm z\le r_0),\qquad
 f_E(z)\le C_0e^{-c_0\norm z},
\end{equation}
where $c_0,C_0,r_0$ are fixed positive constants. Given arbitrarily labeled samples and accuracy $\xi>0$, a polynomial-time algorithm returns a polynomial list containing a homogeneous halfspace with error at most $C\eta+\xi$, with any prescribed high confidence, whenever a homogeneous comparator has error $\eta$. The polynomial is joint in $d,1/\xi$, and log inverse failure probability. Constants may also be retained.
\end{lemma}
\begin{proof}
Encode labels by $\widetilde Y=2Y-1$. For $w\ne0$ define the bounded, scale-invariant sigmoid objective
\begin{equation}\label{eq:sigmoid}
 F_\sigma(w)=\E\left[\frac{1}{1+\exp(\widetilde Y w^TX/(\sigma\norm w))}\right].
\end{equation}
We use the sigmoid surrogate introduced in \cite{dktz20}. Its structural gradient bound extends to~\eqref{eq:wellbehaved}, as the following calculation shows. In the plane of $w$ and a unit comparator normal $u$, choose coordinates with $w=e_2$ and $u=\sin\theta\,e_1+\cos\theta\,e_2$, where $0<\theta<\pi$. Write
\[
 a(t)=\frac{e^{-|t|}}{(1+e^{-|t|})^2},\qquad
 Y^*=\operatorname{sign}(u^TX),
\]
and let $G_\sigma$ denote the objective with the clean label $Y^*$. Then
\[
 -e_1^T\nabla G_\sigma(e_2)
 =\frac1\sigma\E\bigl[Y^*X_1a(X_2/\sigma)\bigr].
\]
Suppose $\sin\theta\ge A\sigma$ for a sufficiently large fixed $A$, and $\sigma$ is sufficiently small. In the two rectangles
\[
 r_0/4\le|x_1|\le r_0/2,\qquad |x_2|\le c\sigma,
\]
choose the fixed $c>0$ small enough that $Y^*x_1=|x_1|$ and the rectangles lie in the core ball of~\eqref{eq:wellbehaved}. The density lower bound and $a(x_2/\sigma)\ge c'$ give a positive contribution bounded below by a fixed constant. The negative contribution can occur only where $|x_1|\le |x_2||\cot\theta|$. Using $f_E\le C_0$ there bounds its magnitude by
\[
 \frac{C}{\sigma\sin^2\theta}
 \int_{\R}x_2^2a(x_2/\sigma)\,dx_2
 \le \frac{C\sigma^2}{\sin^2\theta}.
\]
Consequently,
\[
 -e_1^T\nabla G_\sigma(e_2)
 \ge c-C\frac{\sigma^2}{\sin^2\theta}.
\]
The corrupted labels change this gradient component by at most $C\sqrt{\eta/\sigma}$. Indeed, Cauchy--Schwarz bounds their contribution by a constant multiple of
\[
 \sqrt{\eta}\left(\E\frac{e^{-2|X_2|/\sigma}X_1^2}{\sigma^2}\right)^{1/2}
 \le C\sqrt{\eta/\sigma},
\]
where the final inequality follows by integrating the exponential planar density envelope. Choose $A$ large and then $\kappa$ small. Whenever $\sin\theta\ge A\sigma$ and $\eta\le\kappa\sigma$, the population gradient has a fixed positive lower bound. Adjusting $A$ to convert the sine bound to an angle bound, there are fixed constants $A,\kappa,\rho,\sigma_0>0$ such that
\begin{equation}\label{eq:stationary}
 \norm w=1,\quad\eta\le\kappa\sigma,\quad0<\sigma\le\sigma_0,\quad
 \norm{\nabla F_\sigma(w)}\le\rho
 \quad\Longrightarrow\quad
 \min\{\angle(w,u),\angle(-w,u)\}\le A\sigma.
\end{equation}
Randomized labels are allowed: their conditional corruption probability replaces the indicator in this calculation, and its square is at most itself.

The planar upper envelope also bounds disagreement of two homogeneous halfspaces by $C$ times their angle. Hence~\eqref{eq:stationary} gives a useful classifier by retaining both signs. Both estimates hold under the density assumptions in~\eqref{eq:wellbehaved}.

We next verify that stochastic gradient descent finds an iterate with a sufficiently small population gradient. The envelope implies $\E XX^T\preceq CI$. For a unit $w$, let $g$ be the stochastic gradient of the summand in~\eqref{eq:sigmoid}. Differentiating the normalized margin gives
\begin{equation}\label{eq:sgdbounds}
 w^Tg=0,\qquad \E\norm g^2\le Cd/\sigma^2,
 \qquad \norm{\nabla^2F_\sigma(v)}_{\rm op}\le C/\sigma^2
 \quad(\norm v\ge1).
\end{equation}
The Hessian estimate follows directionally from $\E XX^T\preceq CI$ and the first two derivative bounds of the sigmoid; it has no factor $d$. For the update
\[
 v=w-sg,\qquad w^+=v/\norm v,
\]
tangency ensures that the entire segment $w-tsg$, $0\le t\le1$, has norm at least one. Positive scale invariance gives $F_\sigma(w^+)=F_\sigma(v)$. Smooth descent and conditional expectation therefore yield
\[
 \E[F_\sigma(w^+)\mid w]
 \le F_\sigma(w)-s\norm{\nabla F_\sigma(w)}^2
       +Cs^2d/\sigma^4.
\]
Take $s=c\sigma^4/d$ with a sufficiently small fixed $c$, and $T=C'd/\sigma^4$ with a sufficiently large fixed $C'$. Since $0\le F_\sigma\le1$, the expected average squared gradient can be made less than $\rho^2/4$. Markov's inequality gives, with constant probability, an iterate satisfying~\eqref{eq:stationary}. Keep all iterates and their negatives; independent repeats amplify confidence without estimating any population gradient.

Finally enumerate a geometric grid of $\sigma$ from a sufficiently small fixed multiple of $\xi$ up to $\sigma_0$. If $\eta$ is below a fixed constant, some grid point satisfies $\eta\le\kappa\sigma$ and $\sigma\le C(\eta+\xi)$. If $\eta$ is above that constant, a constant hypothesis already meets a sufficiently large constant-factor guarantee. Rescale the internal $\xi$ to obtain the displayed additive term. The grid, trajectory lengths, repetitions, and resulting list are polynomial.
\end{proof}

\section{Balanced affine learning}\label{sec:balanced}
\begin{proposition}[Balanced affine primitive]\label{prop:balanced}
Fix positive constants $b_0,c_0,C_0$. Let $R$ be log-concave with
\[
 \norm{\E_RX}\le C_0,\qquad c_0I\preceq\Cov_RX\preceq C_0I.
\]
If an affine comparator $H$ has $R(H)\in[b_0,1-b_0]$ and label error $\eta_R$, a sample-only algorithm returns a polynomial list containing an affine classifier of error at most $C\eta_R+\xi$, with prescribed high confidence and joint polynomial complexity in $d,1/\xi$, and log inverse failure probability.
\end{proposition}
\begin{proof}
Assume $d\ge2$; one-dimensional threshold ERM handles $d=1$. If $\eta_R$ is larger than a sufficiently small fixed constant, constants suffice. Otherwise, the observed label-conditional laws are within $O(\eta_R/b_0)$ total variation of the clean laws $R\mid H$ and $R\mid H^c$. Both clean conditional covariances are comparable to $I$, and their mean norms are bounded, by Lemma~\ref{lem:restriction} after a fixed-condition-number affine change.

Lemma~\ref{lem:robust} estimates the two clean means $\mu_+,\mu_-$ to an arbitrarily small fixed Euclidean error $r_0$. Their signed distances from the target boundary are bounded below by a fixed constant. Indeed their nonnegative scalar residuals have density bounded by a constant $B$, so
\[
 \E V\ge\int_0^{1/B}(1-Bt)\,dt=1/(2B).
\]
Thus sufficiently accurate estimates $a_+,a_-$ remain on opposite sides. Repeat these paired estimation attempts independently and retain each resulting segment; with high confidence at least one pair is simultaneously accurate. Norm and segment-length caps discard inadmissible outputs without affecting that pair.

Fix a successful pair and any two-dimensional projection $E$. Each clean class density is bounded below on a fixed-radius ball about $E\mu_+$ or $E\mu_-$. The $R$ marginal dominates each by its fixed class probability. Small mean errors shrink these balls by at most a fixed amount. Log-concavity then gives a uniform lower density bound on equal-radius balls centered at every point of $[Ea_-,Ea_+]$: if $z=(1-t)Ea_-+tEa_++u$, use the same displacement $u$ in the two endpoint balls. The $R$ marginal also has a universal exponential envelope, and every point on the segment has bounded norm. Consequently, for every $a\in[a_-,a_+]$, the translated law of $X-a$ obeys~\eqref{eq:wellbehaved}.

The segment crosses the true boundary. Its length is bounded by a fixed constant, so a grid of Euclidean mesh $\tau$ has $O(1/\tau)$ points and contains one point within distance $\tau$ of the crossing. The homogeneous halfspace through this point with the comparator's normal differs from $H$ only in a strip of width at most $\tau$. The scalar density bound gives disagreement at most $C\tau$. Lemma~\ref{lem:homogeneous} at each grid point, with internal accuracy $c\xi$ and $\tau=c\xi$, supplies a candidate of error at most $C\eta_R+\xi$. Retain all candidates. The list remains polynomial under confidence amplification.
\end{proof}

\section{Local-to-global disagreement for a minority cap}\label{sec:transfer}
Although $K$ need not contain all of $H$, sufficiently small disagreement on $K$ determines the global disagreement up to a factor proportional to $P(H)$. This allows a halfspace learned on $K$ to be used globally.

\begin{lemma}[Parameter-to-global cap bound]\label{lem:global}
In exact cap coordinates~\eqref{eq:coordinates}, write $H=\{n^TZ+a\ge0\}$, where $\norm n=1$ and $1\le a\le A_0$. Let $h=\{m^TZ+b\ge0\}$, $\norm m=1$. There are universal $\beta_0,C>0$ such that
\begin{equation}\label{eq:global}
 \beta:=\norm{m-n}+|b-a|\le\beta_0
 \quad\Longrightarrow\quad
 P(H\triangle h)\le Cp\beta.
\end{equation}
\end{lemma}
\begin{proof}
Project onto the plane of $m,n$, completing it arbitrarily if the normals are parallel. The projected law of $Q$ is isotropic log-concave. Use coordinates $z=(x,t)$ with $n=e_1$ and $H=\{x\ge-a\}$. Let $q$ be this projected conditional density and $f$ the original projected density divided by $p$. Because the projection contains $n$, conditioning commutes with the projection: $f=q$ on $H$. In particular,
\[
 q(z)\le Ae^{-c\norm z},\qquad q(0)\ge c.
\]
Disagreement implies
\[
 |x+a|\le |(m-n)^Tz+b-a|
 \le\beta(1+\norm z)
 \le\beta(1+a+|t|+|x+a|).
\]
Absorbing the last term shows that it lies in the strip
\begin{equation}\label{eq:strip}
 |x+a|\le C\beta(1+|t|).
\end{equation}
Inside $H$ the envelope for $q$ bounds $f$ by $Ce^{-c|t|}$.

For a point $z=(x,t)$ in the strip outside $H$, put $\lambda=a/(-x)\in(0,1)$ and $y=\lambda z$. Then $y$ is on the cap boundary. Concavity of $\log f$ on the segment between $0$ and $z$ gives
\begin{align*}
 f(z)&\le q(y)^{1/\lambda}q(0)^{-(1-\lambda)/\lambda}\\
 &\le\exp\{C/\lambda-c\norm z\}.
\end{align*}
Here $\norm y/\lambda=\norm z$. On~\eqref{eq:strip},
\[
 1/\lambda=\frac{-x}{a}\le1+C\beta(1+|t|).
\]
Taking $\beta_0$ sufficiently small absorbs the resulting positive multiple of $\beta|t|$ into the negative tail exponent. Thus $f(z)\le Ce^{-c'|t|}$ throughout the strip, on either side. Integrating its width proves
\[
 \int_{H\triangle h}f\le C\beta\int_\R(1+|t|)e^{-c'|t|}\,dt\le C\beta.
\]
Multiplying by $p$ gives~\eqref{eq:global}.

This proof is valid for compact supports. If $f(z)>0$ and $f(0)>0$, log-concavity forces a positive trace at every interior point of their segment, including $y$. A zero boundary trace therefore cannot hide positive density beyond the boundary. In dimension one the same argument bounds the density on an interval of length $O(\beta)$; no extra projected coordinate is needed.
\end{proof}

\begin{lemma}[Local disagreement controls affine parameters]\label{lem:parameters}
Suppose $R$ is log-concave in the exact cap coordinates, satisfies~\eqref{eq:regioncov}, and gives both classes of $H$ a fixed positive probability. Then for fixed $\delta_0,C>0$, every halfspace $h=\{m^TZ+b\ge0\}$, $\norm m=1$, satisfies
\begin{equation}\label{eq:parameters}
 R(H\triangle h)\le\delta_0
 \quad\Longrightarrow\quad
 \norm{m-n}+|b-a|\le C R(H\triangle h).
\end{equation}
Constants cannot meet this small-disagreement premise.
\end{lemma}
\begin{proof}
Project onto the two normals. The two class-conditional laws under $R$ have bounded means and covariance comparable to $I$, by Lemma~\ref{lem:restriction}. Their projected mean segment crosses the target boundary at a point $z_0$ of bounded norm. As in Proposition~\ref{prop:balanced}, log-concavity gives a fixed radius $r$ and density lower bound $\ell>0$ throughout the ball $B(z_0,r)$.

Translate its center to zero. The target becomes $\{n^Tz\ge0\}$ and the candidate becomes $\{m^Tz+c_h\ge0\}$, where $c_h=m^Tz_0+b$. Let $D_B$ be the unnormalized planar area of their disagreement in this ball. Then $D_B\le R(H\triangle h)/\ell$.

The central target bisects the ball. The difference between the candidate's area and half the ball's area is at most $D_B$. Parallel chords of a circular ball give a lower bound $c\min\{r^2,r|c_h|\}$ on that area difference. For sufficiently small $D_B$, this yields $|c_h|\le CD_B$. The candidate and its same-normal central halfplane disagree in area at most $C|c_h|$. By the triangle inequality, the two central halfplanes then disagree in area at most $CD_B$. Their disagreement area is exactly a fixed multiple of their oriented angle, giving $\norm{m-n}\le CD_B$ and excluding reversed orientation. Finally, $n^Tz_0=-a$ and
\[
 |b-a|\le|m^Tz_0+b|+\norm{m-n}\norm{z_0}\le CD_B.
\]
This proves~\eqref{eq:parameters}. In dimension one use the corresponding interval-length argument. A constant disagrees with $H$ on at least the smaller class probability, so it is excluded after reducing $\delta_0$.
\end{proof}

\begin{theorem}[Minority-mass transfer]\label{thm:transfer}
Let $K$ and $R=P\mid K$ be as in Proposition~\ref{prop:balance}. There are fixed $\delta_0,c,C>0$ such that every affine classifier $h$ with $R(H\triangle h)\le\delta_0$ obeys
\begin{equation}\label{eq:transfer}
 cp\,R(H\triangle h)\le P(H\triangle h)\le Cp\,R(H\triangle h).
\end{equation}
\end{theorem}
\begin{proof}
Reduce $\delta_0$ until Lemma~\ref{lem:parameters} puts the parameters within the range of Lemma~\ref{lem:global}; combining the lemmas proves the upper bound. For the lower bound, $P(H\triangle h)\ge P((H\triangle h)\cap K)=P(K)R(H\triangle h)$, and Proposition~\ref{prop:balance} gives $P(K)\ge K_0p/2$.
\end{proof}

The small-error premise is essential. An always-positive classifier has conditional disagreement bounded away from zero but may have global disagreement close to one when $p$ is small.

\section{The learning algorithm}\label{sec:algorithm}
We describe the algorithm and prove its error guarantee; Section~\ref{sec:implementation} establishes the sample and running-time bounds. The constants are chosen in the following order. Take $B=2$ and choose the robust-estimation tolerance $r$ small enough for Lemma~\ref{lem:hinge} and the central-halfspace argument in Proposition~\ref{prop:balance}. Fix the hinge-loss bound $C_1$, then choose $K_0$ with fixed slack beyond $4C_1$. This determines the balanced-learning constants and the transfer threshold $\delta_0$. Finally choose the noise-ratio threshold $\alpha$ and the internal accuracy multiplier sufficiently small. All these choices are independent of $d,p,\varepsilon,\delta$.

Fix a comparator $H$ with error $\eta$. Orient its minority class as positive, so $p=P(H)\le1/2$; the algorithm runs both label orientations and includes both constants. For sufficiently small fixed $\alpha>0$, if $\eta\ge\alpha p$, the negative constant has error at most
\begin{equation}\label{eq:constant}
 p+\eta\le(1+1/\alpha)\eta.
\end{equation}
If $p\le c\varepsilon$, it has error at most $\eta+c\varepsilon$. Consequently the only difficult regime is
\begin{equation}\label{eq:difficult}
 p\ge c\varepsilon,\qquad \eta/p\le\alpha.
\end{equation}
Choose internal accuracy $\varepsilon_0=c_*\varepsilon$, where $c_*$ is fixed sufficiently small after all risk constants are fixed.

Let $q=\Pp(Y=1)$ in the chosen orientation. Then $|q-p|\le\eta$. Estimate it by $\widehat q$ to additive $c\varepsilon_0$. In the difficult regime it is a fixed-factor proxy for $p$, and in fact can be made arbitrarily close in relative constant terms. Branches with $\widehat q$ below a fixed multiple of $\varepsilon_0$ need only constants. Choose a fixed $K_0$ as in Proposition~\ref{prop:balance}, large enough for the fixed-factor use of $q$, and a cutoff
\[
 q_*=(100K_0)^{-1}.
\]
If $\widehat q\ge q_*$, run Proposition~\ref{prop:balanced} directly under the original isotropic law, retaining all its candidates. In a difficult branch $p$ is then bounded below by a fixed constant, so both comparator classes are balanced. No restriction is needed in that branch.

If $\widehat q<q_*$, estimate the moments of the observed positive law. Its total variation distance from $Q=P\mid H$ is $O(\eta/p)$. For example, comparing the unnormalized measures $\Pp(X\in\cdot,Y=1)$ and $\Pp(X\in\cdot,H)$, their variation discrepancy is at most $\eta$, and normalizing costs only a constant multiple of $1/p$ under~\eqref{eq:difficult}. An iid maximal coupling gives iid clean $Q$ samples with iid mismatch indicators of this probability. A Chernoff bound yields the replacement model in Lemma~\ref{lem:robust} when $\alpha$ is small enough. The learner uses only the observed samples; the coupling verifies the contamination assumption of the robust estimator.

Independently repeat the robust estimator and retain all admissible coordinate systems. With high confidence one satisfies~\eqref{eq:robust}. In each system optimize the unlabeled hinge to additive $c\widehat q$, and use an empirical quantile to construct $K$ of mass approximately $K_0\widehat q$, containing its central set $S$. In the successful branch Proposition~\ref{prop:balance} applies with universal constants. The fixed cutoff ensures $P(K)<1/2$. Rejection sampling supplies labeled examples under $R=P\mid K$, where the conditional noise satisfies
\begin{equation}\label{eq:noiseamp}
 \eta_R=\Pp(Y\ne H\mid K)\le\eta/P(K)\le C\eta/p.
\end{equation}
Run Proposition~\ref{prop:balanced} in the estimated coordinates $\widehat Z$, with accuracy $\varepsilon_0$. On a successful estimation attempt, $Z=A\widehat Z+e$ from Lemma~\ref{lem:hinge} has singular values bounded above and below by universal constants and $\norm e\le r$. Thus~\eqref{eq:regioncov} also gives bounded mean and comparable covariance for $R$ in the estimated coordinates, as required by that proposition. Pull each candidate back to the original coordinates. One candidate obeys
\[
 R(H\triangle h)\le\err_{D\mid K}(h)+\eta_R
 \le C\eta/p+C\varepsilon_0.
\]
Choose $\alpha$ and the multiplier in $\varepsilon_0$ so that this is below the $\delta_0$ in Theorem~\ref{thm:transfer}. Then
\begin{equation}\label{eq:cancel}
 P(H\triangle h)\le Cp(\eta/p+\varepsilon_0)
 \le C\eta+C\varepsilon_0.
\end{equation}
Adding $\eta$ bounds its original label error. The same affine $h$ is used on all of $\R^d$. Retain every candidate from every surviving attempt, and select the lowest empirical original-distribution error on a fresh validation sample. This adds at most $C\varepsilon_0$ with high probability.

The candidate list includes the outputs of all retained coordinate systems, mean estimates, translations, and optimization runs. Selection uses a fresh labeled validation sample from the original distribution. For the proof, fix any comparator within an arbitrarily small additive tolerance of $\OPT_D$ before the algorithm's randomness. Its successful branch exists by the argument above. Taking the comparator tolerance and internal accuracy sufficiently small proves Theorem~\ref{thm:main}, once the polynomial implementation below is established.

\section{Polynomial implementation and confidence}\label{sec:implementation}
We specify polynomial sample and iteration bounds for every stage, including attempts whose statistical estimates are inaccurate. All raw sample batches are independent of prior outputs; accepted conditional samples depend on the previously chosen restriction. A conditional sample request is implemented by a bounded number of ordinary iid draws and rejection using the explicit halfspace.

\subsection{Bounds on valid coordinates}
\begin{lemma}[A priori cap moment bounds]\label{lem:capbounds}
If the original $P$ is isotropic log-concave and $P(H)=p$, then
\begin{equation}\label{eq:capbounds}
 \lambda_{\min}(\Sigma_Q)\ge cp^2,
 \quad \lambda_{\max}(\Sigma_Q)\le C\log^2(e/p),
 \quad \norm{\mu_Q}\le C\log(e/p).
\end{equation}
\end{lemma}
\begin{proof}
Every scalar marginal of $P$ has density at most a universal constant. Its density conditional on $H$ is at most $C/p$, so the density-to-variance bound from Lemma~\ref{lem:restriction} gives the first inequality. For every unit $v$, the original directional tail obeys $P(|v^TX|>t)\le Ae^{-at}$. Thus
\[
 \E[(v^TX)^2\mid H]
 \le\int_0^\infty 2t\min\{1,Ae^{-at}/p\}\,dt
 \le C\log^2(e/p).
\]
Taking directional suprema gives the covariance upper bound and the mean norm bound.
\end{proof}

Whenever $p\ge c\varepsilon_0$, a valid robust output lies in explicit polynomial spectral and norm bounds. Reject all outputs outside slightly looser bounds, including nonpositive-definite matrices. On every surviving branch,
\[
 \E_P\norm{\widehat\Sigma^{-1/2}(X-\widehat\mu)}^2\le M_2
\]
for a known polynomial $M_2=\poly(d,1/\varepsilon_0)$. This follows directly from isotropy and the imposed bounds, even when the output is inaccurate. All eigendecompositions and inverses here are in the declared real-arithmetic model.

\subsection{Sample optimization of the hinge}
Fix a surviving coordinate map and let $Z$ denote its transformed feature. Choose a deterministic radius $R_0$ polynomial in $M_2$ and $1/\varepsilon_0$. Truncate only the objective:
\[
 L_{R_0}(v)=\E[(1+v^TZ)_+\ind_{\{\norm Z\le R_0\}}],\qquad\norm v\le B.
\]
The population marginal itself is not truncated. Uniformly over feasible $v$,
\begin{equation}\label{eq:hingetail}
 0\le L(v)-L_{R_0}(v)
 \le M_2/R_0^2+BM_2/R_0.
\end{equation}
Take $R_0$ large enough that the right side is at most $c\varepsilon_0$, hence at most a sufficiently small multiple of $p$ in the difficult branch. The truncated summands are bounded by $1+BR_0$ and are $R_0$-Lipschitz in $v$. A Euclidean net of the radius-$B$ ball with mesh $\zeta/R_0$, a Hoeffding bound, and Lipschitz extension show that uniform empirical error at most $\zeta$ holds with
\[
 n=O\!\left(\frac{(1+BR_0)^2}{\zeta^2}
       \left[d\log\frac{CBR_0}{\zeta}+\log\frac1{\delta'}\right]\right)
\]
samples. Here $\zeta=c\varepsilon_0$ or a small multiple of $\widehat q$, both polynomially bounded below in surviving relevant branches. The net is for the proof only. Projected subgradient descent on the empirical convex objective obtains additive $\zeta$ optimization error in $O(B^2R_0^2/\zeta^2)$ steps. Evaluating a full empirical subgradient per step still costs a polynomial. Combining optimization, uniform convergence, and~\eqref{eq:hingetail} gives the desired additive $O(p)$ hinge solution. Abort a zero slope in the rare branch; it cannot occur in a valid branch because its loss would be one.

\subsection{Quantiles and rejection}
Use a fresh sample and an empirical upper-tail quantile of $v^TZ$ with target mass $k=K_0\widehat q$. The uniform empirical CDF error can be made at most $e=c\varepsilon_0$ by a polynomial sample bound, with confidence $1-\delta'$. Include the order-statistic discretization error in $e$. Define the final threshold as the minimum of this empirical threshold and $-1/2$. This choice ensures $K\supseteq S$. In a valid branch the true mass of $S$ is substantially below $k$, so this minimum does not change the expanded empirical quantile. Its true mass is $k\pm O(e)=\Theta(p)$.

Before substantial rejection sampling, estimate acceptance on another fresh batch. Abort a branch whose estimate is below a fixed polynomial lower threshold. Put a deterministic polynomial cap on every rejection count; Chernoff bounds show that all valid branches with mass $\Theta(p)\ge c\varepsilon_0$ survive with the allocated confidence. The same safeguards apply to observed positive samples and the two label classes in balanced restrictions. Rejection changes the distribution only by selecting accepted iid points; no synthetic or conditional oracle is used.

\subsection{Balanced lists and optimization budgets}
For each balanced-primitive call, independently repeat paired robust class-mean estimation. Retain every pair inside fixed norm and segment-length caps, and run all its $O(1/\varepsilon_0)$ grid points. A successful pair lies inside these caps by the proved conditional moment bounds. Inaccurate mean estimates may add candidates to the list, which are evaluated on the independent validation sample.

Every homogeneous run has a fixed geometric bandwidth grid and a predetermined polynomial trajectory length. Even when the statistical estimates are inaccurate, the sigmoid's stochastic gradient is finite for every finite sample, the initial vector is unit, and the unnormalized update has norm at least one, since $\norm{w-sg}^2=1+s^2\norm g^2$. Thus the retraction is defined, and the predetermined iteration budget bounds the operation count without assuming that branch's geometry. All retained hypotheses are affine functions of the original features.

There is a predetermined polynomial upper bound $N$ on the number of branches and output candidates. Allocate statistical failure probability $\delta'=\delta/(CN)$ to each fresh concentration or rejection guarantee. Robust moments need only constant probability per attempt; $O(\log(N/\delta))$ independent attempts leave at least one good output with high confidence. In balanced calls repeat the paired attempts, so both mean estimates are simultaneously good in at least one retained pair. Homogeneous trajectories are repeated similarly. The products of these fixed-stage polynomial list sizes remain polynomial. Every attempt terminates within its prescribed budget.

\subsection{Sigmoid evaluation using arithmetic alone}\label{sec:numerical}
The sigmoid in Section~\ref{sec:homogeneous} need not be an exact elementary-function oracle. Let $N_{\max}$ be the deterministic polynomial bound on the number of original draws in the complete capped algorithm. From the isotropic coordinate tails and a union bound, a deterministic
\[
 B_{\rm raw}=C\sqrt d\log(CdN_{\max}/\delta)
\]
ensures that every original draw has norm at most $B_{\rm raw}$ except with probability at most $\delta/10$. The implemented algorithm aborts if this bound is violated. All surviving coordinate maps, class-mean outputs, and translations already have polynomial caps; hence every feature supplied to a homogeneous run then has norm at most a deterministic polynomial $M$.

The sigmoid derivative uses $a(t)=e^{-|t|}/(1+e^{-|t|})^2$. To approximate its stochastic gradient to norm error at most a small fixed $\zeta>0$, it suffices to approximate $a$ uniformly to error
\[
 e_{\rm num}=c\zeta\sigma/(M+1).
\]
Find the smallest nonnegative integer $m$ with $2^{-m}\le e_{\rm num}/2$ by repeated halving. For $|t|>m$, return zero: $a(t)\le e^{-|t|}\le e^{-m}\le2^{-m}$. On $[-m,m]$, the Taylor polynomial for $e^{-|t|}$ of degree $Cm$ has absolute error at most $c e_{\rm num}$ for a sufficiently large fixed $C$, by the exponential-series remainder bound. Clip its value to $[0,1]$ and apply $s\mapsto s/(1+s)^2$, whose derivative is bounded on that interval. Since $m=O(\log(2/e_{\rm num}))$, this requires polynomially many arithmetic operations and comparisons. Norms and normalization use square roots as permitted by the computational model.

Keep the approximate gradient as a scalar multiple of the exact tangent projection $X-(w^TX)w$. Tangency, and thus the scale-invariant retraction argument, remains exact. To justify the analysis without conditioning the data on a norm event, consider a fictitious oracle: it uses this approximation on bounded draws and the exact gradient on oversized draws. Its pointwise vector error is at most $\zeta$ everywhere. It is tangent, and its conditional bias has norm at most $\zeta$. The descent inequality consequently gains terms at most
\[
 s\zeta\norm{\nabla F_\sigma(w)}+Cs^2\zeta^2/\sigma^2.
\]
Young's inequality absorbs the first into a fixed fraction of the gradient descent term plus $Cs\zeta^2$. Choosing $\zeta$ sufficiently small relative to the fixed stationarity threshold, and adjusting the fixed step and iteration constants, proves the same list guarantee. The implemented and fictitious algorithms coincide unless an oversized original draw occurs, whose probability is at most $\delta/10$.  Since $\sigma\ge c\varepsilon_0$, the arithmetic evaluation cost stays polynomial.

\subsection{Validation and constants}
A fresh original-distribution sample of size
\[
 O\bigl(\varepsilon_0^{-2}(\log N+\log(1/\delta))\bigr)
\]
uniformly estimates the errors of all retained candidates to additive $\varepsilon_0$. Select an empirical minimizer. Combining~\eqref{eq:constant}, the balanced branch, and~\eqref{eq:cancel} proves the theorem after the fixed rescaling of $\varepsilon_0$.

In dimension one, threshold empirical risk minimization in both orientations, including constants, has the stronger additive agnostic guarantee and polynomial complexity. If the convention for ``proper halfspace'' excludes constants, replace them by $\ind\{X_1\ge T\}$ and $\ind\{X_1\ge-T\}$ for $T=C\log(1/\varepsilon_0)$, choosing $C$ using the universal isotropic log-concave tail bound. These finite thresholds differ from the constants on probability at most $\varepsilon_0$, which is absorbed by the choice of $\varepsilon_0$. The comparator infimum is unchanged by allowing constants, since their errors are limits of finite thresholds.

\section{Unknown affine normalization and confidence amplification}
\label{sec:affine-extension}
We now remove the isotropy assumption and obtain the sample bound in Corollary~\ref{cor:general}. The preprocessing uses only empirical moments of the marginal. Labels are ignored in this step.

\begin{lemma}[Stability under approximate normalization]\label{lem:approx-normalization}
Fix constants $M,c_-,c_+>0$. The guarantee of Theorem~\ref{thm:main} continues to hold when its isotropy assumption is replaced by
\[
 \norm{\E X}\le M,\qquad c_-I\preceq\Cov(X)\preceq c_+I.
\]
The risk constant and the coefficients of the polynomial complexity bound may depend on $M,c_-,c_+$.
\end{lemma}
\begin{proof}
All cap geometry in Sections~\ref{sec:cap} and~\ref{sec:transfer} is expressed in the cap's own mean and covariance, so these arguments are unchanged. In the branch with a comparator of constant minority mass, the displayed assumptions are precisely those of Proposition~\ref{prop:balanced}.

For the remaining implementation bounds, every unit projection of $X$ has a density bounded by a fixed constant and a tail bounded by $A\exp(-at)$, with constants depending only on $M,c_-,c_+$. The proof of Lemma~\ref{lem:capbounds} therefore gives
\[
 \lambda_{\min}(\Sigma_Q)\ge cp^2,\quad
 \lambda_{\max}(\Sigma_Q)\le C\log^2(e/p),\quad
 \norm{\mu_Q}\le C\log(e/p).
\]
Also $\E\norm X^2\le C(d+1)$, which supplies the bound $M_2$ used in the hinge optimization. The same directional tails give the sample-norm bound in Section~\ref{sec:numerical}. These are the only uses of isotropy outside the cap normalization. Adjusting the fixed constants in the spectral, norm, and iteration bounds proves the claim.
\end{proof}

\begin{lemma}[Empirical whitening at constant confidence]\label{lem:whitening}
Let $X$ have a full-dimensional log-concave law with unknown mean $\mu$ and covariance $\Sigma$. For $n=C d^2$ independent samples, let
\[
 \widehat\mu=\frac1n\sum_{i=1}^nX_i,
 \qquad
 \widehat\Sigma=\frac1n\sum_{i=1}^n
 (X_i-\widehat\mu)(X_i-\widehat\mu)^T.
\]
For a sufficiently large universal $C$, with probability at least $9/10$,
\begin{equation}\label{eq:empirical-whitening}
 \norm{\Sigma^{-1/2}(\widehat\mu-\mu)}\le\tfrac18,
 \qquad
 \tfrac12\Sigma\preceq\widehat\Sigma\preceq\tfrac32\Sigma.
\end{equation}
On this event, the transformed feature $W=\widehat\Sigma^{-1/2}(X-\widehat\mu)$ satisfies
\[
 \norm{\E W}\le1,\qquad
 \tfrac23 I\preceq\Cov(W)\preceq2I.
\]
\end{lemma}
\begin{proof}
Use the true standardized variables $Z_i=\Sigma^{-1/2}(X_i-\mu)$ for the analysis. Their one-dimensional fourth moments are bounded by a universal constant. Cauchy--Schwarz therefore gives
\[
 \E\norm Z^4
 =\sum_{j,k}\E[Z_j^2Z_k^2]\le Cd^2.
\]
Writing $S=n^{-1}\sum_iZ_iZ_i^T$ and $\overline Z=n^{-1}\sum_iZ_i$, independence yields
\[
 \E\norm{S-I}_F^2
 =\frac1n\E\norm{ZZ^T-I}_F^2\le\frac{Cd^2}{n},
 \qquad
 \E\norm{\overline Z}^2=\frac dn.
\]
For a sufficiently large constant in $n=Cd^2$, Markov's inequality and a union bound imply
\[
 \norm{S-I}_{\rm op}\le\tfrac18,
 \qquad \norm{\overline Z}\le\tfrac18
\]
with probability at least $9/10$. Since
\[
 \Sigma^{-1/2}\widehat\Sigma\Sigma^{-1/2}
 =S-\overline Z\,\overline Z^T,
\]
these estimates imply~\eqref{eq:empirical-whitening}. Congruence by $\widehat\Sigma^{-1/2}$ gives the covariance bounds for $W$, and
\[
 \norm{\E W}^2
 =(\widehat\mu-\mu)^T\widehat\Sigma^{-1}(\widehat\mu-\mu)
 \le2\norm{\Sigma^{-1/2}(\widehat\mu-\mu)}^2\le1.
\]
The algorithm computes the empirical quantities directly in the original coordinates. The unknown normalization is used only to establish their accuracy.
\end{proof}

\begin{proof}[Proof of Corollary~\ref{cor:general}]
First construct a learner with a fixed success probability. Estimate $\mu$ and $\Sigma$ as in Lemma~\ref{lem:whitening}, and check that $\widehat\Sigma$ is positive definite. If it is, apply the learner of Lemma~\ref{lem:approx-normalization}, with internal accuracy $\varepsilon/4$ and failure probability $1/10$, to fresh samples of $(W,Y)$. Conditional on the whitening event, its assumptions hold with the fixed bounds $M=1$, $c_-=2/3$, and $c_+=2$. An invertible affine change preserves the class of affine halfspaces, their classification errors, and $\OPT_D$.

If the learner returns a nonconstant halfspace $\ind\{v^TW\ge t\}$, pull it back as
\begin{equation}\label{eq:pullback}
 w=\widehat\Sigma^{-1/2}v,
 \qquad b=t+w^T\widehat\mu.
\end{equation}
If it returns a constant, replace that constant in $W$ coordinates by
\[
 h_0(W)=\ind\{W_1\ge T\},
 \qquad h_1(W)=\ind\{W_1\ge-T\},
 \qquad T=C\log(C/\varepsilon),
\]
respectively. On the whitening event the one-dimensional log-concave tail bounds imply
\[
 \Pp(h_0(W)\ne0)+\Pp(h_1(W)\ne1)\le\varepsilon/4
\]
after choosing $C$ sufficiently large. Their pullbacks have nonzero normals and finite thresholds. To choose a threshold using arithmetic alone, find the least integer $j\ge1$ for which $2^j\ge C/\varepsilon$ by repeated doubling, and use $T=C'j$ with a sufficiently large fixed $C'$. Each successful run thus returns a nonconstant halfspace with error at most $C\OPT_D+\varepsilon/2$.

If the empirical covariance is singular, or a subsequent stage exceeds its prescribed budget, return the fixed halfspace $\ind\{x_1\ge0\}$. No absolute spectral bound is imposed on the original $\widehat\Sigma$: its eigenvalues may reflect an arbitrarily ill-conditioned marginal. The norm and spectral bounds of Section~\ref{sec:implementation} are applied after whitening, in $W$ coordinates. Consequently every run has a fixed polynomial sample and arithmetic budget, including those for which~\eqref{eq:empirical-whitening} fails. On the successful whitening event, all probabilistic guarantees of the normalized learner apply. A union bound shows that one full run succeeds with probability at least $4/5$.

Repeat this entire procedure independently
\[
 J=\left\lceil C\log(2/\delta)\right\rceil
\]
times and retain its $J$ outputs. With probability at least $1-\delta/2$, one has error at most $C\OPT_D+\varepsilon/2$. Use an independent validation sample of size
\[
 m=O\!\left(\varepsilon^{-2}
       \bigl(\log(2J)+\log(2/\delta)\bigr)\right)
\]
to estimate all $J$ errors to additive $\varepsilon/4$, simultaneously with probability at least $1-\delta/2$. Their empirical minimizer has true error at most the smallest candidate error plus $\varepsilon/2$. This proves the risk bound.

Each fixed-confidence run uses $P_0(d,\varepsilon^{-1})$ samples for a fixed polynomial $P_0$. Since $\log(2J)=O(\log(2/\delta))$, the total is at most $P(d,\varepsilon^{-1})\log(2/\delta)$ for a fixed polynomial $P$. Computing the empirical moments, transforming samples, and evaluating the $J$ candidates on the common validation sample all take polynomially many real-arithmetic operations. None of these operation counts depends on the eigenvalue ratio of $\Sigma$.
\end{proof}

\section{Discussion}
The minority-mass comparison~\eqref{eq:transfer} explains why learning on a balanced restriction can yield a global constant-factor agnostic guarantee. Conditioning amplifies the label error by a factor of order $1/p$, while the comparison between conditional and global disagreement contributes the compensating factor $p$. Robust cap normalization and the zonoid-depth objective make this construction accessible from labeled samples. The output remains a single affine halfspace.

Empirical whitening extends the result to unknown full-dimensional log-concave marginals, and independent repetition makes the sample complexity linear in $\log(2/\delta)$. The polynomial exponents arising from robust moment estimation and the search over translations are not optimized. Obtaining smaller exponents, extending the geometry beyond log-concave marginals, and establishing finite-bit complexity bounds are natural directions for further work.

\end{document}